\documentclass[dvipsnames,format=sigconf,anonymous=false,review=false,nonacm=true]{acmart}
\usepackage{colortbl}
\usepackage{multirow}
\usepackage{multicol}
\usepackage{booktabs} 
\usepackage[ruled,linesnumbered]{algorithm2e}
\usepackage{color}
\AtBeginDocument{%
  }

\setcopyright{acmlicensed}
\copyrightyear{2018}
\acmYear{2018}
\acmDOI{XXXXXXX.XXXXXXX}
\acmConference[Conference acronym 'XX]{Make sure to enter the correct
  conference title from your rights confirmation email}{June 03--05,
  2018}{Woodstock, NY}
\acmISBN{978-1-4503-XXXX-X/2018/06}

\begin{document}

\title{Accurate Peak Detection in Multimodal Optimization via Approximated Landscape Learning}

\author{Zeyuan Ma}
\authornote{Both authors contributed equally to this research.}
\email{scut.crazynicolas@gmail.com}
\orcid{0000-0001-6216-9379}
\affiliation{%
  \institution{South China University of Technology}
  \city{Guangzhou}
  \state{Guangdong}
  \country{China}
}

\author{Hongqiao Lian}
\authornotemark[1]
\email{qiao2471035068@163.com}
\orcid{0009-0005-4424-6189}
\affiliation{%
  \institution{South China University of Technology}
  \city{Guangzhou}
  \state{Guangdong}
  \country{China}
}

\author{Wenjie Qiu}
\email{wukongqwj@gmail.com}
\orcid{0009-0000-1965-8863}
\affiliation{%
  \institution{South China University of Technology}
  \city{Guangzhou}
  \state{Guangdong}
  \country{China}
}

\author{Yue-Jiao Gong}
\authornote{Corresponding Author}
\email{gongyuejiao@gmail.com}
\orcid{0000-0002-5648-1160}
\affiliation{%
  \institution{South China University of Technology}
  \city{Guangzhou}
  \state{Guangdong}
  \country{China}}

\renewcommand{\shortauthors}{Trovato et al.}

\begin{abstract}
Detecting potential optimal peak areas and locating the accurate peaks in these areas are two major challenges in Multimodal Optimization problems (MMOPs). To address them, much efforts have been spent on developing novel searching operators, niching strategies and multi-objective problem transformation pipelines. Though promising, existing approaches more or less overlook the potential usage of landscape knowledge. In this paper, we propose a novel optimization framework tailored for MMOPs, termed as APDMMO, which facilitates peak detection via fully leveraging the landscape knowledge and hence capable of providing strong optimization performance on MMOPs. Specifically, we first design a novel surrogate landscape model which ensembles a group of non-linear activation units to improve the regression accuracy on diverse MMOPs. Then we propose a free-of-trial peak detection method which efficiently locates potential peak areas through back-propagation on the learned surrogate landscape model. Based on the detected peak areas, we employ SEP-CMAES for local search within these areas in parallel to further improve the accuracy of the found optima. Extensive benchmarking results demonstrate that APDMMO outperforms several up-to-date baselines. Further ablation studies verify the effectiveness of the proposed novel designs. The source-code is available at ~\href{}{https://github.com/GMC-DRL/APDMMO}. 
\end{abstract}




\maketitle

\section{INTRODUCTION}
Problems with multiple global optima are known as multimodal optimization problems (MMOPs) and are frequently encountered in real-world tasks such as electromagnetic machine design~\cite{electromagnetic} and protein structure prediction~\cite{protein}.  Locating all optima, or as many as possible, can assist decision-makers in changing policies to adapt to dynamic environments, or provide comprehensive knowledge of the problem to analysts. Due to these advantages, research on MMOPs has gained popularity continuously in recent years. 

Evolutionary computation (EC) algorithms act as meta-heuristics with high efficiency and robustness and show promising optimization performance in many optimization fields. ~However, some early stage EC algorithms are designed for optimizing unimodal problems and cannot be utilized to optimize multimodal problems directly due to the genetic drift and selection pressure. To maintain the diversity of the population during the optimization, researchers make efforts in several following methods: 1) The most popular technique, named niching, is used to divide the whole population into several small sub-populations, each responsible for the sub-space around it, enhancing the exploration of the solution space. Classic niching methods include crowding~\cite{crowding}, speciation~\cite{speciation}, fitness sharing~\cite{sharing} and restricted tournament selection (RTS)~\cite{RTS}. 2) New evolutionary operators are designed to improve the performance of algorithms, such as enhancing the optimization around each possible optimal solution~\cite{FBK-DE} or expanding mutation operators with historical individuals~\cite{history}. 3) Transforming MMOPs into multi-objective optimization problems by introducing diversity and quality as conflicting objectives to balance exploration and exploitation~\cite{MOMMOP,EMOMMO}. However, as an important guiding information: local/global landscape structures of MMOPs are less discussed than the above aspects. If it is aware of the landscape details of an MMOP, one can avoid redundant searches for the same region based on landscape knowledge~\cite{LADE}, or describe the shape of peaks through clustering~\cite{hillvallea} or multimodal transformation~\cite{EMOMMO}. Nevertheless, such simple landscape analysis only cover the in-time optimization information: current and historical population. This is insufficient to profile comprehensive landscape information of an MMOP, especially those in unexplored areas.  

In this paper, we propose \textbf{\underline{A}}ccurate \textbf{\underline{P}}eak \textbf{\underline{D}}etection in \textbf{\underline{M}}ulti\textbf{\underline{m}}odal \textbf{\underline{O}}ptimization via Approximated Landscape Learning (APDMMO), a novel multimodal optimization framework that fully utilizes landscape of an MMOP to locate all of its optima. Specifically, APDMMO includes three sequential stages. In the first surrogate learning stage, we design a novel neural network architecture, termed as Landscape Learner, which ensembles diverse non-linear activation units for accurate surrogate learning. By uniformly sampling moderate sample points from the decision space, we train the Landscape Learner effectively as a surrogate model of the target MMOP. In the second peak detection stage, with the trained surrogate model, we employ an adamW~\cite{adamw} optimizer with a multi-start strategy to grid search potential peak areas through efficient gradient descent along the surrogate model. Note that this stage will incur no function evaluations because we use the model instead of the objective function for evaluation. After using a clustering method to identify the general positions of peaks represented by converged points, the task in the third parallel local search stage is to refine each peak to obtain the actual global optima. Due to the strong ability in local searching, SEP-CMAES~\cite{sep-cma-es} is adopted as the optimizer to search around each peak, and the results will be stored in an archive as the found optima. The experimental results show that APDMMO achieves superior performance against several up-to-date baselines tailored for MMOPs. Now we summarize our contributions:

\begin{itemize}
\item In this paper, we propose APDMMO, a three-stage optimization framework that utilizes sufficient landscape information of MMOPs and a combination of gradient-based and gradient-free optimizers to achieve state-of-the-art results on CEC 2013 MMOP benchmark.

\item To achieve such promising performance, we first design a novel network architecture to ensemble diverse non-linear activation units for accurate surrogate learning. Assisted by the learned surrogate, we further propose an efficient multi-start adamW strategy for fast peak detection. Once potential peak areas are determined, we employ multiple SEP-CMAES instances in these areas for further local search and solution accuracy refinement.
\item Through extensive benchmarking, we demonstrate the effectiveness and superiority of APDMMO compared with $16$ up-to-date baselines, covering all primary development trends of addressing MMOPs. Furthermore, a series of fruitful observations and corresponding discussions are provided in our ablation studies, which underscore the important roles of the core designs in APDMMO.

\end{itemize}

\begin{figure*}[t]
  \centering
  \includegraphics[width=0.7\textwidth]{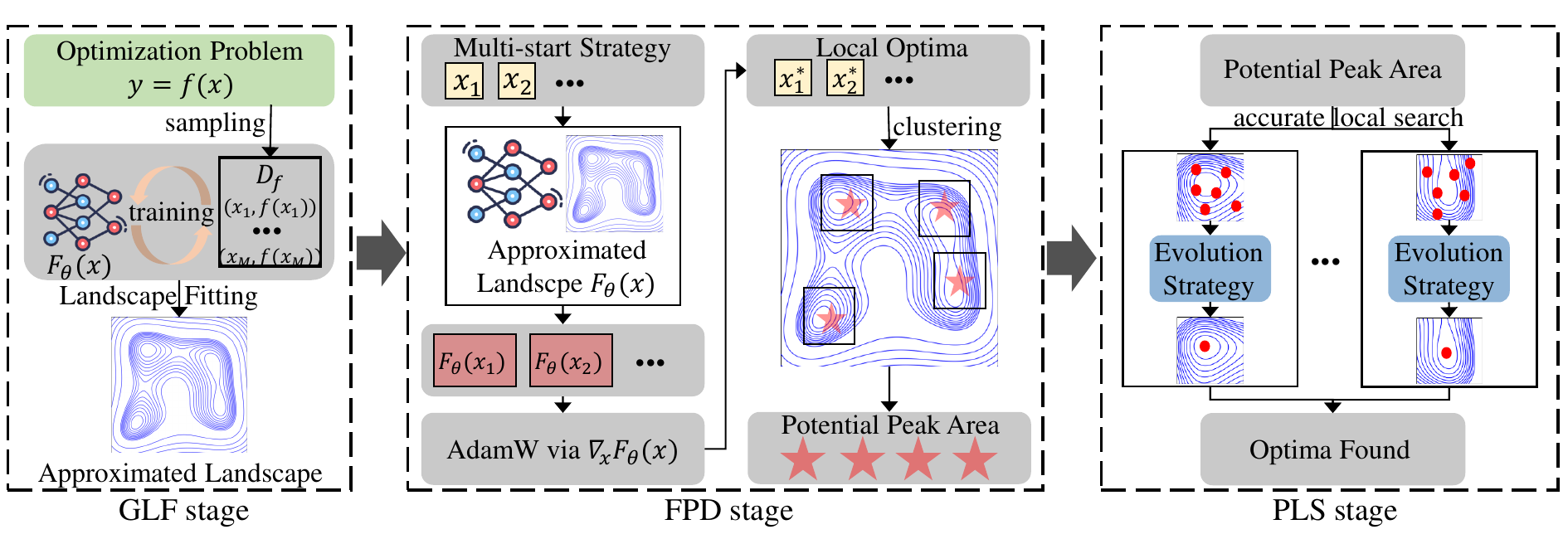}
  \caption{The overall workflow of APDMMO, with three sequential stages, from left to right.}
  \label{fig:overview}
\end{figure*}

\section{RELATED WORKS}

\subsection{Evolutionary Multimodal Optimization}\label{sec:2.1}

In recent years, researchers have incorporated EC algorithms with various techniques to enhance the performance for MMOPs, which can be roughly classified into three aspects.

 \subsubsection{Niching Technique}
The first improvement involves integrating niching techniques with conventional evolutionary algorithms, dividing the population into multiple subpopulations to locate different optima. Thomsen et al.~\cite{CDE} extended Differential Evolution (DE) with the crowding technique, and each offspring competed with its nearest parent. Qu et al.~\cite{qu2012differential} limited each individual to evolve towards the nearest optimal solution. Jiang et al.~\cite{jiang2021optimizing} utilized an internal Genetic Algorithm (GA) to determine the niching centers. To reduce the number of parameters, Zhang et al.~\cite{zhang2019parameter} introduced a Voronoi neighborhood, while Wang et al.~\cite{wang2019automatic} introduced Affinity Propagation Clustering as a niching method. To enhance the performance of particle swarm optimization (PSO), Li et al.~\cite{li2009niching} made each particle  interact with its immediate neighbors based on a ring topology. Fieldsend et al.~\cite{fieldsend2014running} proposed a niching migratory multi-swarm optimizer to dynamically manage swarms. In addition, algorithms integrating niching techniques with GA~\cite{farias2018parent}, evolution strategy (ES)~\cite{ahrari2017multimodal} or estimation of distribution algorithms (EDAs)~\cite{yang2016multimodal}  also achieve good results in MMOPs.


\subsubsection{New Evolutionary Operator}
The second aspect of improvement designs new evolutionary operators to enhance local search and maintain diversity during optimization. In particular, Agrawal et al.~\cite{agrawal2022solving} proposed a novel mutation strategy that generates four virtual random individuals to calculate the differential vectors.  Liao et al.~\cite{history} operated a new mutation strategy to select random individuals from the combination of a history archive and the current population. Biswas et al.~\cite{biswas2014improved} proposed an DE variant with a parent-centric mutation and crowding technique. Lin et al.~\cite{FBK-DE} introduced two new mutation operators based on the key-points within clusters, while Sun~\cite{sun2023differential} proposed two new operators to refine the exploration area of individuals. Zhao et al.~\cite{zhao2019local} proposed a mutation strategy of global and niching interaction to guide an individual to a more promising position. 

\subsubsection{Multi-objective Problem Transformation}
Some researches consider constructing two or more objectives to transform MMOPs into multi-objective optimization problems (MOOPs), in which Pareto optimal solutions correspond to the peaks in MMOPs. Basak et al.~\cite{basak2012multimodal} conducted a bi-objective DE with objectives about the multimodal function and the mean distance of a solution from all other members, while Wang et al.~\cite{MOMMOP} proposed a novel method to construct two conflicting objectives to process the transformation. Cheng et al.~\cite{EMOMMO} transformed MMOPs into MOOPs by adding a diversity objective, and then constructed the peak detection and local search after obtaining the Pareto optimal solutions.

Although existing works have improved the performance of EC algorithms in MMOPs, few algorithms consider utilizing landscape information. Lin et al.~\cite{LADE} utilized landscape knowledge to guide exploration and initialization, while Cheng et al.~\cite{EMOMMO} or Mareeet al.~\cite{hillvallea} approximated the landscape through MOOPs transformation or clustering. However, such landscape analyses rely on the historical or current population, which is insufficient to provide comprehensive knowledge of the whole space.

\subsection{Learning for EC}
In recent years, learning for EC has become a popular topic because it can offer increased flexibility and expandability to EC algorithms. Two notable branches in this field are meta-Black-Box Optimization (MetaBBO)~\cite{ma2024toward,ma2024metabox} and surrogate-assisted EAs (SAEA)~\cite{jin2011surrogate}. 

\subsubsection{MetaBBO}
MetaBBO approaches facilitate a bi-level optimization scheme to meta-learn a parameterized control policy in a data-driven way. This method configures the low-level black-box optimizer during optimization to reduce the design efforts needed for unseen tasks. The policy can be learned in various ways (e.g., supervised learning~\cite{gomes2021meta,chen2017learning,wu2023decn}, reinforcement learning~\cite{sun2021learning,sharma2019deep,ma2024neural,guo2024configx}, in-context learning~\cite{ma2024llamoco}). MetaBBO with reinforcement learning (MetaBBO-RL) involves fine-tuning as a Markov Decision Process (MDP) and trains an agent to suggest EC-related operators based on the status. As representative works, Ma et al.~\cite{ma2024auto} proposed an RL-based framework to configure the exploration-exploitation tradeoff  automatically and dynamically. Guo et al.~\cite{guo2024deep} leveraged an RL-based agent to determine suitable algorithms at each optimization interval. Chen et al.~\cite{chen2024symbol} developed a symbolic equation generator to generate symbolic update equations for the optimization process. In another aspect, MetaBBO with supervised learning (MetaBBO-SL) parameterizes the operators as models and learns the operations directly. For example, Li et al. utilized deep neural networks to parameterize various evolutionary modules in GA~\cite{li2023b2opt} or DE~\cite{li2024glhf}, and verified their convergence and generalization.

\subsubsection{SAEA}

For high-dimensional or expensive optimization problems, SAEA employs surrogates (approximation model)  to replace function evaluations, thereby reducing the search time of evolutionary optimization algorithms. It is commonly employed in expensive multi-objective optimization problems~\cite{song2021kriging,knowles2006parego}, expensive constrained optimization problems~\cite{wang2018global,regis2013evolutionary}, and expensive global optimization problems~\cite{li2021three}. Existing works in this field can be categorized into three main groups: global SAEA, local SAEA and global-local-collaborative SAEA. Sun et al.~\cite{sun2017surrogate} proposed to combine a global surrogate-assisted social learning-based PSO for exploring the global space with a surrogate-assisted PSO to search for global optimum. Ong et al.~\cite{ong2003evolutionary} developed local surrogate models for each individual and its neighbors to search better offspring by optimizing the surrogates. Wang et al.~\cite{wang2019novel} used a global surrogate model to prescreen the objective function values of offspring and then constructed a local surrogate model with the current best solutions to assist in finding the optimum.

Some existing studies have introduced learning methods for MMOPs. Lian et al.~\cite{lian2024rlemmo} designed a quality and diversity related reward scheme to train an RL agent for the selection of mutation operators. Hong et al.~\cite{hong2024reinforcement} proposed an RL-based neighborhood range selection strategy to adjust the size of subpopulation. Ji et al.~\cite{ji2021dual} combines a global surrogate and local surrogates to simultaneously explore and exploit multiple modalities. Du et al.~\cite{du2024surrogate} used a global surrogate to predict modalities and a joint surrogate-assisted local search to exploit the optimum. In APDMMO, we utilize an artificial neural network to approximate the landscape of the problem, assisting in detecting peaks.

\section{Methodology}
Our proposed APDMMO, whose overall framework is illustrated in Figure~\ref{fig:overview}, comprises three sequential stages: Global Landscape Fitting~(GLF) stage, Free-of-trial Peak Detection~(FPD) stage and Parallel Local Search~(PLS) stage. In GLF stage, for a given target optimization problem, APDMMO first fully utilizes the landscape information of the target problem, by consuming a portion of function evaluations to construct a surrogate model which we term it as landscape learner $F_\theta$. In FPD stage, we directly employ gradient descent on multiple randomly sampled start points using the trained landscape learner $F_\theta$, which does not consume function evaluations and shows fast convergence toward several potential peak areas. In PLS stage, we employ the Evolution Strategy as the local optimizer to locate optima in these potential peak areas in parallel. In the rest of this section, we first elaborate technical details of GLF stage, FPD stage and PLS stage and motivations behind in Section~\ref{sec4.1}, Section~\ref{sec4.2} and Section~\ref{sec4.3} respectively. 

\begin{figure}[t]
  \centering
  \includegraphics[width=0.9\columnwidth]{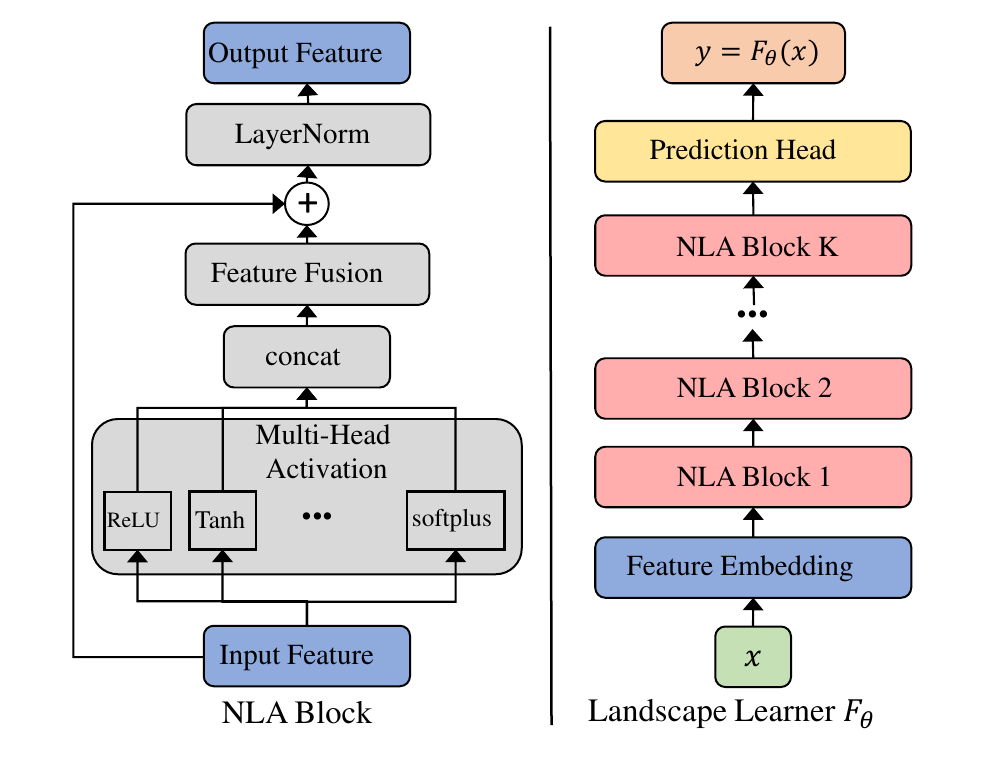}
  \caption{Architecture design of the Landscape Learner.}
  \label{fig:network}
  \vspace{-3mm}
\end{figure}

\subsection{Global Landscape Fitting}\label{sec4.1}
As we aforementioned in the end of Section~\ref{sec:2.1}, although some existing methods\cite{EMOMMO,hillvallea} address MMOPs by approximating the landscape of local searching areas during the optimization progress, they unavoidably overlook the importance of the global landscape structures that might be very useful for peak detection in MMOPs. In APDMMO, we design a Global Landscape Fitting~(GLF) stage to learn global landscape information of MMOPs and we show how such information be helpful for the peak detection stage in Section~\ref{sec4.2}. Before that, we elaborate how GLF works as follows. 

\subsubsection{Landscape Learner} 
Using Artificial Neural Network~(ANN) as a surrogate model to assist mathematical modeling and optimization has been proven effective, e.g., in aerodynamic design~\cite{surrogate-review-aerodynamic} and comparing black box optimization problems~\cite{saleem2022using}. Recent studies also indicate that using an ensemble of different ANN models might be more effective to profile both the global landscape structure and the multiple local modalities~(peaks) in MMOPs~\cite{surrogate-emsemble}. Following this idea, we design a simple yet effective neural network architecture, termed as Landscape Learner, whose structure is illustrated in Figure~\ref{fig:network}. Instead of using ensemble of different ANN models, we propose a novel non-linear activation~(NLA) Block design, which uses an ensemble of $22$ non-linear activation functions to adapt for complex and diverse local landscape properties in MMOPs, as shown in the left of Figure~\ref{fig:network}. Specifically, given an input feature $S_{in} \in \mathbb{R}^{h}$, $h$ is the hidden dimension of the Landscape Learner, it is activated by the $22$ activation functions in the Multi-Head Activation module to attain multiple non-linear post-activation features. Then these features are concatenated into a hidden feature vector $S_{hidden}\in \mathbb{R}^{22\times h}$. To attain the output feature $S_{out}^h$, the hidden feature goes through a linear mapping-based Feature Fusion layer which maps $S_{hidden}$ to $\mathbb{R}^h$, and a Layer Normalization~\cite{layernorm} layer which normalizes the feature value to avoid gradient explosion. Note that we additionally add a skip connection between input and hidden feature, which analogs the residual module in the Transformer to enhance the learning effectiveness. The $22$ activation functions are \emph{ELU}, \emph{Hardshrink}, \emph{Hardsigmoid}, \emph{Hardtanh}, \emph{Hardswish}, \emph{LeakyReLU}, \emph{LogSigmoid}, \emph{PReLU}, \emph{ReLU}, \emph{ReLU6}, \emph{RReLU}, \emph{SELU}, \emph{CELU}, \emph{GELU}, \emph{Sigmoid}, \emph{SiLU}, \emph{Mish}, \emph{Softplus}, \emph{Softshrink}, \emph{Softsign}, \emph{Tanh}, \emph{Tanhshrink}. They provide sufficient non-linearity for Landscape Learner to approximate the landscapes of MMOPs.

Now we summarize the complete workflow of the Landscape Learner, which is shown in the right of Figure~\ref{fig:network}. Given the input $x$ as a $d$-dimensional sample point of an MMOP, it is first linearly embedded by a Feature Embedding layer as the input feature $S_{in}$. Then we stack $K$ NLA blocks to improve the expressiveness of the model. The output feature of $K$-th NLA block is fed into a Prediction Head, a linear mapping  that maps the output feature to a numerical value to serve as the prediction of the objective value of $x$. In the rest of this paper, we use $F_\theta$ to refer to the Landscape Learner.

\begin{algorithm}[t]
\caption{The Training Process of Landscape Learner.}
\label{alg:model}\small
\KwIn{Objective function $f$, Dataset $D_f:\{X_f,Y_f\}$, Initialized model $F_\theta$, the number of objective value levels $s$, Batch size $N_{batch}$, Training epochs $T_{mse}$, Learning rate $\eta$.}
\KwOut{Optimal landscape model $F_{\theta^*}$.}
\For{$epoch \leftarrow 1$ \KwTo $T_{mse}$}{
    \For{each batch in $D_f$}{
        Use EPM sampling strategy to select $N_{batch}$ samples from $D_f$, with number of levels $s$.\\
        $\mathcal{L}(\theta) \leftarrow \frac{1}{N_{batch}}\sum_{i=1}^{N_{batch}} MSE(f(x_i), F_\theta(x_i))$; \\
        $\theta \leftarrow \theta - \eta \nabla_\theta \mathcal{L}(\theta)$; 
    }
}
\KwRet{The trained $F_\theta$}
\end{algorithm}

\subsubsection{Training Workflow}
Generally speaking, the training workflow of GLF resembles the classic regression, where given an MMOP $f$  and a corresponding dataset $D_f: \{X_f,Y_f=f(X_f)\}$ which comprises $M$ sample points on $f$ and their objective values, the objective of GLF is to minimize the regression loss as below:
\begin{equation}
    \mathcal{L}(\theta) = \frac{1}{M}\sum_{i=1}^{M} MSE(f(x_i),F_\theta(x_i))
\end{equation}
where $MSE(\cdot,\cdot)$ denotes mean square error function. We present the training process of $\theta$ in Algorithm~\ref{alg:model}. The only difference of this training process compared to classic regression is that we adopt example-proportional mixing~(EPM) sampling strategy~\cite{distribution1} (line $3$) during training to improve the training effectiveness. Specifically, when we sample a mini-batch of sample points, we first divide all sample points in $D_f$ into $s$ levels according to their objective values. Then we sample same number of sample points from these levels to construct the mini-batch used for training. The motivation of the EPM sampling strategy is to address the unbalanced distribution in $D_f$. Since we construct the dataset $D_f$ by uniform sampling on the searching space, the sampled points in $D_f$ is in an unbalanced distribution considering their objective values, due to the complex landscape of $f$. With the EPM sampling strategy, we re-balance the sample points with different objective value levels, ensuring that they are averagely fitted so the accuracy of the model is improved. 

\begin{algorithm}[t]
\caption{Workflow of APDMMO.}
\label{alg:workflow}\small
\KwIn{MMOP $f$, Maximum function evaluations $maxFes$, Ratio of function evaluations $r$ for GLF; Number of start points $N_{fpd}$ , step size $\eta_{fpd}$, gradient decent steps $G_{fpd}$, DBSCAN's $eps$ and $Minpts$ for FPD stage; SEP-CMAES's variation $\sigma$ and optimization generation $G_{pls}$, population size $n_{pls}$, early stop threshold $\epsilon_{es}$ and $k_{es}$}
\KwOut{Optima set $\mathbb{O}$.}
\state{/*  GLF stage */\\}
Construct $D_f : \{X_f, f(X_f)\}$ with size $r \times maxFes$;\\
Train Landscape Learner $F_{\theta}$ following Algorithm~\ref{alg:model};\\

\state{/* FPD stage */\\}
$X\leftarrow$ Sample $N_{fpd}$ points uniformly;\\
\For{$step \leftarrow 1$ \KwTo $G_{fpd}$}{
        $X\leftarrow X - \eta_{fpd} \nabla_{X} F_{\theta}(X)$;\\
}
$\{c_1,c_2,...\} \leftarrow $ DBSCAN$(X;eps, Minpts)$;\\
Find $p_i^*  \in c_i$ and add it to Archive $P$;\\ 
Sort $P$ in ascending order of $F_{\theta}(P)$;\\
\state{/* PLS stage */\\}
$fes \leftarrow 0$;\\
$k \leftarrow 0$;\\
\While{$fes < (1 - r) \times maxFes$}{
        $\mathbb{O}_k\leftarrow $ SEP-CMAES$(P_k, \sigma, G_{pls}, n_{pls}, \epsilon_{es}, k_{es})$;\\
        $\mathbb{O} \leftarrow \mathbb{O} \cup \mathbb{O}_k$;\\
        Increase $fes$ by the consumed function evaluations;\\
        $k \leftarrow (k + 1) \% |P|$;\\
}

\KwRet{Optima set $\mathbb{O}$}
\end{algorithm}

\subsection{Free-of-trial Peak Detection}\label{sec4.2}
The second stage in APDMMO is the Free-of-trial Peak Detection~(FPD) stage. Within this stage, we detect potential peak areas of an MMOP $f$ based on the Landscape Learner $F_\theta$ trained in previous GLF stage and advanced clustering techniques. We first employ an adamW optimizer with multi-start strategy to locate potential peaks, through gradient descent on the approximated landscape $F_\theta$~(left part in the middle box of Figure~\ref{fig:overview}). Specifically, suppose $F_\theta$ is trained as an accurate surrogate model of $f$, then it is expected to profile the global and local landscape structures of $f$ comprehensively. Since $F_\theta$ is differentiable, given a start point $x^0$ in $f$'s searching space, we can leverage this convenience to search potential peaks in landscape of $F_\theta$ by iterative gradient descent:
\begin{equation}
    x^{t+1} = x^t - \eta_{fpd} \nabla_x F_\theta(x)
\end{equation}
where $\eta_{fpd}$ is a constant step size. We iteratively call this gradient descent procedure for $G_{fpd}$ steps to attain a potential peak $x^*$. A key advantage in FPD stage is that: we can leverage multi-start strategy to uniformly sample massive numbers of start points, say $N_{fpd}$ such start points, and search the corresponding potential peaks following the gradient descent procedure in parallel on GPU efficiently. More importantly, this procedure does not consume any actual function evaluations of the MMOP $f$ itself.

After the above gradient descent procedure, we attain a collection of potential peaks. One key insight is that these potential peaks are only optima or local optima on $F_\theta$, which might be inequivalent to those on the original MMOP $f$ since there is always prediction bias in $F_\theta$. However, a biased prediction of optima is more or less useful information for us to locate potential peak areas. Based on this insight, we leverage DBSCAN~\cite{DBSCAN} to separate the found potential peaks into a group of clusters, as shown in the right in the middle box of Figure~\ref{fig:overview}. DBSCAN is an advanced and easy-to-use clustering method that excels in discovering clusters based on the density of the distribution of large-scale point sets, with only two simple control parameters: $\epsilon$~(eps) and $minPts$~(minimum number of points required to form a dense region). For each cluster $c_i$ we obtain from DBSCAN, we identify the cluster center as the potential peak $p^*_i$ in it with the lowest $F_\theta(x)$ and add it into a sorted archive $\mathbb{P}:\{p_1^*,p_2^*,...\}$, which we use for parallel local search stage.

\subsection{Parallel Local Search}\label{sec4.3}
The final stage in APDMMO is the Parallel Local Search~(PLS) stage, where for each cluster center $p_i^*$, we employ an SEP-CMAES~\cite{ros2008simple} optimizer with $p_i^*$ as initial mean vector and a constant variation $\sigma$ to further improve the solution precision in that cluster. After $G_{pls}$ generations of optimization, the actual optima~(peaks) of the MMOP $f$ are found, with high precision. If the improvement in objective values does not exceed the early stop threshold $\epsilon_{es}$ in successive $k_{es}$ evaluations, the current SEP-CMAES process will be considered stagnant and terminate early. Note that, in PLS stage we optimize those cluster centers using the MMOP $f$ as the evaluation function, hence this stage consumes function evaluations ($FEs$).     

We now summarize the overall workflow of APDMMO. The pseudocode of APDMMO is presented in Algorithm~\ref{alg:workflow}. Given an MMOP $f$ and corresponding maximum function evaluations~($maxFEs$) can be used for optimizing it, we use a portion of $r\times maxFEs$ function evaluations to construct a training dataset $D_f$ (see section~\ref{res-ablation} for the ablation discussion of $r$) and use it to train the Landscape Learner $F_\theta$~(line $1\sim 3$, GLF stage). We then use the learned $F_\theta$ and DBSCAN clustering method to detect potential peak areas and attain a sorted Archive $P$~(line $4 \sim 11$, FPD stage). At last, we run SEP-CMAES in each potential peak area in $P$ to locate the actual optima of $f$~(line $12\sim19$). We have to note that APDMMO relies on the cooperation of FPD stage and PLS stage. On the one hand, the FPD stage provides economic and effective peak detection results without requiring function evaluations. On the other hand, the PLS stage provides robust refinement on approximated peak areas and locate actual optima efficiently. We provide fruitful experimental results in the next section to verify the performance of APDMMO, as well as the effectiveness of core designs.

\begin{table*}[ht]
    \caption{Comprehensive comparison results on CEC2013 MMOP benchmark of APDMMO and baselines.}
    \label{main_exp}
    \tiny
    \resizebox{0.85\textwidth}{!}{
    \begin{tabular}{c|c|c|c|c|c|c|c|c|c}
    \hline
    problem(dimension) & APDMMO & RLEMMO & NDC-DE & NetCDE & ESPDE & PMODE&NBNC-PSO-ES&FBK-DE&DIDE \\ \hline
    F1(1D)& \cellcolor{gray!30}\textbf{1.000}(1.000)&\cellcolor{gray!30}\textbf{1.000}(1.000)&\cellcolor{gray!30}\textbf{1.000}(1.000)&\cellcolor{gray!30}\textbf{1.000}(1.000)&\cellcolor{gray!30}\textbf{1.000}(1.000)&\cellcolor{gray!30}\textbf{1.000}(1.000)&\cellcolor{gray!30}\textbf{1.000}(1.000)&\cellcolor{gray!30}\textbf{1.000}(1.000)&\cellcolor{gray!30}\textbf{1.000}(1.000)\\ \hline
F2(1D)& \cellcolor{gray!30}\textbf{1.000}(1.000)&\cellcolor{gray!30}\textbf{1.000}(1.000)&\cellcolor{gray!30}\textbf{1.000}(1.000)&\cellcolor{gray!30}\textbf{1.000}(1.000)&\cellcolor{gray!30}\textbf{1.000}(1.000)&\cellcolor{gray!30}\textbf{1.000}(1.000)&\cellcolor{gray!30}\textbf{1.000}(1.000)&\cellcolor{gray!30}\textbf{1.000}(1.000)&\cellcolor{gray!30}\textbf{1.000}(1.000)\\ \hline
F3(1D)& \cellcolor{gray!30}\textbf{1.000}(1.000)&\cellcolor{gray!30}\textbf{1.000}(1.000)&\cellcolor{gray!30}\textbf{1.000}(1.000)&\cellcolor{gray!30}\textbf{1.000}(1.000)&\cellcolor{gray!30}\textbf{1.000}(1.000)&\cellcolor{gray!30}\textbf{1.000}(1.000)&\cellcolor{gray!30}\textbf{1.000}(1.000)&\cellcolor{gray!30}\textbf{1.000}(1.000)&\cellcolor{gray!30}\textbf{1.000}(1.000)\\ \hline
F4(2D)& \cellcolor{gray!30}\textbf{1.000}(1.000)&\cellcolor{gray!30}\textbf{1.000}(1.000)&\cellcolor{gray!30}\textbf{1.000}(1.000)&\cellcolor{gray!30}\textbf{1.000}(1.000)&\cellcolor{gray!30}\textbf{1.000}(1.000)&\cellcolor{gray!30}\textbf{1.000}(1.000)&\cellcolor{gray!30}\textbf{1.000}(1.000)&\cellcolor{gray!30}\textbf{1.000}(1.000)&\cellcolor{gray!30}\textbf{1.000}(1.000)\\ \hline
F5(2D)& \cellcolor{gray!30}\textbf{1.000}(1.000)&\cellcolor{gray!30}\textbf{1.000}(1.000)&\cellcolor{gray!30}\textbf{1.000}(1.000)&\cellcolor{gray!30}\textbf{1.000}(1.000)&\cellcolor{gray!30}\textbf{1.000}(1.000)&\cellcolor{gray!30}\textbf{1.000}(1.000)&\cellcolor{gray!30}\textbf{1.000}(1.000)&\cellcolor{gray!30}\textbf{1.000}(1.000)&\cellcolor{gray!30}\textbf{1.000}(1.000)\\ \hline
F6(2D)& \cellcolor{gray!30}\textbf{1.000}(1.000)&\cellcolor{gray!30}\textbf{1.000}(1.000)&0.999(0.960)&\cellcolor{gray!30}\textbf{1.000}(1.000)&\cellcolor{gray!30}\textbf{1.000}(1.000)&\cellcolor{gray!30}\textbf{1.000}(1.000)&\cellcolor{gray!30}\textbf{1.000}(1.000)&0.990(0.820)&\cellcolor{gray!30}\textbf{1.000}(1.000)\\ \hline
F7(2D)& \cellcolor{gray!30}\textbf{1.000}(1.000)&0.925(0.040)&0.881(0.000)&0.947(0.255)&0.963(0.360)&0.672(0.000)&0.967(0.140)&0.813(0.000)&0.921(0.040)\\ \hline
F8(3D)&0.916(0.000)&0.824(0.000)&0.941(0.000)&0.999(0.902)&0.880(0.000)&0.616(0.000)&0.808(0.000)&0.824(0.000)&0.692(0.000)\\ \hline
F9(3D)& \cellcolor{gray!30}\textbf{1.000}(1.000)&0.610(0.000)&0.458(0.000)&0.511(0.000)&0.729(0.000)&0.324(0.000)&0.540(0.000)&0.425(0.000)&0.571(0.000)\\ \hline
F10(2D)& \cellcolor{gray!30}\textbf{1.000}(1.000)&\cellcolor{gray!30}\textbf{1.000}(1.000)&\cellcolor{gray!30}\textbf{1.000}(1.000)&\cellcolor{gray!30}\textbf{1.000}(1.000)&\cellcolor{gray!30}\textbf{1.000}(1.000)&\cellcolor{gray!30}\textbf{1.000}(1.000)&\cellcolor{gray!30}\textbf{1.000}(1.000)&\cellcolor{gray!30}\textbf{1.000}(1.000)&\cellcolor{gray!30}\textbf{1.000}(1.000)\\ \hline
F11(2D)& \cellcolor{gray!30}\textbf{1.000}(1.000)&0.950(0.720)&\cellcolor{gray!30}\textbf{1.000}(1.000)&0.984(0.902)&\cellcolor{gray!30}\textbf{1.000}(1.000)&\cellcolor{gray!30}\textbf{1.000}(1.000)&\cellcolor{gray!30}\textbf{1.000}(1.000)&\cellcolor{gray!30}\textbf{1.000}(1.000)&\cellcolor{gray!30}\textbf{1.000}(1.000)\\ \hline
F12(2D)& \cellcolor{gray!30}\textbf{1.000}(1.000)&0.995(0.960)&0.941(0.520)&0.904(0.471)&0.930(0.560)&\cellcolor{gray!30}\textbf{1.000}(1.000)&\cellcolor{gray!30}\textbf{1.000}(1.000)&0.935(0.480)&\cellcolor{gray!30}\textbf{1.000}(1.000)\\ \hline
F13(2D)& \cellcolor{gray!30}\textbf{1.000}(1.000)&0.793(0.000)&\cellcolor{gray!30}\textbf{1.000}(1.000)&0.667(0.000)&0.793(0.080)&0.953(0.720)&\cellcolor{gray!30}\textbf{1.000}(1.000)&\cellcolor{gray!30}\textbf{1.000}(1.000)&0.987(0.920)\\ \hline
F14(3D)& \cellcolor{gray!30}\textbf{1.000}(1.000)&0.667(0.000)&0.833(0.160)&0.667(0.000)&0.727(0.000)&0.800(0.000)&0.847(0.100)&0.901(0.460)&0.773(0.020)\\ \hline
F15(3D)&0.750(0.000)&0.748(0.000)& \cellcolor{gray!30}\textbf{0.753}(0.000)&0.630(0.000)&0.730(0.000)&0.750(0.000)&0.738(0.000)&0.713(0.000)&0.748(0.000)\\ \hline
F16(5D)& \cellcolor{gray!30}\textbf{0.833}(0.000)&0.667(0.000)&0.667(0.000)&0.667(0.000)&0.667(0.000)&0.667(0.000)&0.723(0.000)&0.707(0.000)&0.667(0.000)\\ \hline
F17(5D)& \cellcolor{gray!30}\textbf{0.750}(0.000)&0.525(0.000)&0.633(0.000)&0.480(0.000)&0.685(0.000)&0.405(0.000)&0.718(0.000)&0.665(0.000)&0.593(0.000)\\ \hline
F18(10D)& \cellcolor{gray!30}\textbf{0.667}(0.000)&0.567(0.000)&\cellcolor{gray!30}\textbf{0.667}(0.000)&\cellcolor{gray!30}\textbf{0.667}(0.000)&0.660(0.000)&0.500(0.000)&\cellcolor{gray!30}\textbf{0.667}(0.000)&\cellcolor{gray!30}\textbf{0.667}(0.000)&\cellcolor{gray!30}\textbf{0.667}(0.000)\\ \hline
F19(10D)& \cellcolor{gray!30}\textbf{0.625}(0.000)&0.341(0.000)&0.412(0.000)&0.461(0.000)&0.445(0.000)&0.245(0.000)&0.538(0.000)&0.520(0.000)&0.543(0.000)\\ \hline
F20(20D)& \cellcolor{gray!30}\textbf{0.500}(0.000)&0.155(0.000)&0.355(0.000)&0.380(0.000)&0.265(0.000)&0.240(0.000)&0.483(0.000)&0.450(0.000)&0.355(0.000)\\ \hline
Rank& \cellcolor{gray!30}\textbf{4.9}&10.425&8.4&10.2&9.25&10.05&6.65&8.425&8.1\\\hline
 \multicolumn{2}{c|}{+  /  =  /  -}&13/7/0&9/9/2&11/8/1&12/8/0&10/10/0&9/11/0&11/9/0&10/10/0\\ \hline

problem(dimension)&APDMMO&VNCDE&CMSA-ES-DIPS&hillvallea&RS-CMSA&EMOMMO&MMOMOP&NMMSO&dADE2	\\ \hline
F1(1D)&\cellcolor{gray!30}\textbf{1.000}(1.000)&\cellcolor{gray!30}\textbf{1.000}(1.000)&\cellcolor{gray!30}\textbf{1.000}(1.000)&\cellcolor{gray!30}\textbf{1.000}(1.000)&\cellcolor{gray!30}\textbf{1.000}(1.000)&\cellcolor{gray!30}\textbf{1.000}(1.000)&\cellcolor{gray!30}\textbf{1.000}(1.000)&\cellcolor{gray!30}\textbf{1.000}(1.000)&\cellcolor{gray!30}\textbf{1.000}(1.000)\\ \hline
F2(1D)&\cellcolor{gray!30}\textbf{1.000}(1.000)&\cellcolor{gray!30}\textbf{1.000}(1.000)&\cellcolor{gray!30}\textbf{1.000}(1.000)&\cellcolor{gray!30}\textbf{1.000}(1.000)&\cellcolor{gray!30}\textbf{1.000}(1.000)&\cellcolor{gray!30}\textbf{1.000}(1.000)&\cellcolor{gray!30}\textbf{1.000}(1.000)&\cellcolor{gray!30}\textbf{1.000}(1.000)&\cellcolor{gray!30}\textbf{1.000}(1.000)\\ \hline
F3(1D)&\cellcolor{gray!30}\textbf{1.000}(1.000)&\cellcolor{gray!30}\textbf{1.000}(1.000)&\cellcolor{gray!30}\textbf{1.000}(1.000)&\cellcolor{gray!30}\textbf{1.000}(1.000)&\cellcolor{gray!30}\textbf{1.000}(1.000)&\cellcolor{gray!30}\textbf{1.000}(1.000)&\cellcolor{gray!30}\textbf{1.000}(1.000)&\cellcolor{gray!30}\textbf{1.000}(1.000)&\cellcolor{gray!30}\textbf{1.000}(1.000)\\ \hline
F4(2D)&\cellcolor{gray!30}\textbf{1.000}(1.000)&\cellcolor{gray!30}\textbf{1.000}(1.000)&\cellcolor{gray!30}\textbf{1.000}(1.000)&\cellcolor{gray!30}\textbf{1.000}(1.000)&\cellcolor{gray!30}\textbf{1.000}(1.000)&\cellcolor{gray!30}\textbf{1.000}(1.000)&\cellcolor{gray!30}\textbf{1.000}(1.000)&\cellcolor{gray!30}\textbf{1.000}(1.000)&\cellcolor{gray!30}\textbf{1.000}(1.000)\\ \hline
F5(2D)&\cellcolor{gray!30}\textbf{1.000}(1.000)&\cellcolor{gray!30}\textbf{1.000}(1.000)&\cellcolor{gray!30}\textbf{1.000}(1.000)&\cellcolor{gray!30}\textbf{1.000}(1.000)&\cellcolor{gray!30}\textbf{1.000}(1.000)&\cellcolor{gray!30}\textbf{1.000}(1.000)&\cellcolor{gray!30}\textbf{1.000}(1.000)&\cellcolor{gray!30}\textbf{1.000}(1.000)&\cellcolor{gray!30}\textbf{1.000}(1.000)\\ \hline
F6(2D)&\cellcolor{gray!30}\textbf{1.000}(1.000)&0.967(0.540)&\cellcolor{gray!30}\textbf{1.000}(1.000)&\cellcolor{gray!30}\textbf{1.000}(1.000)&0.900(0.160)&\cellcolor{gray!30}\textbf{1.000}(1.000)&\cellcolor{gray!30}\textbf{1.000}(1.000)&0.997(0.940)&0.833(0.002)\\ \hline
F7(2D)&\cellcolor{gray!30}\textbf{1.000}(1.000)&0.746(0.000)&0.991(0.800)&\cellcolor{gray!30}\textbf{1.000}(1.000)&0.997(0.900)&\cellcolor{gray!30}\textbf{1.000}(1.000)&\cellcolor{gray!30}\textbf{1.000}(1.000)&\cellcolor{gray!30}\textbf{1.000}(1.000)&0.757(0.000)\\ \hline
F8(3D)&0.916(0.000)&0.558(0.000)&0.594(0.000)&0.920(0.000)&0.582(0.000)& \cellcolor{gray!30}\textbf{1.000}(1.000)&\cellcolor{gray!30}\textbf{1.000}(1.000)&0.981(0.180)&0.660(0.000)\\ \hline
F9(3D)&\cellcolor{gray!30}\textbf{1.000}(1.000)&0.298(0.000)&0.686(0.000)&0.945(0.000)&0.737(0.000)&0.952(0.000)&\cellcolor{gray!30}\textbf{1.000}(1.000)&0.917(0.000)&0.335(0.000)\\ \hline
F10(2D)&\cellcolor{gray!30}\textbf{1.000}(1.000)&\cellcolor{gray!30}\textbf{1.000}(1.000)&\cellcolor{gray!30}\textbf{1.000}(1.000)&\cellcolor{gray!30}\textbf{1.000}(1.000)&0.998(0.980)&\cellcolor{gray!30}\textbf{1.000}(1.000)&\cellcolor{gray!30}\textbf{1.000}(1.000)&\cellcolor{gray!30}\textbf{1.000}(1.000)&\cellcolor{gray!30}\textbf{1.000}(1.000)\\ \hline
F11(2D)&\cellcolor{gray!30}\textbf{1.000}(1.000)&\cellcolor{gray!30}\textbf{1.000}(1.000)&0.987(0.920)&\cellcolor{gray!30}\textbf{1.000}(1.000)&0.960(0.780)&\cellcolor{gray!30}\textbf{1.000}(1.000)&0.716(0.020)&\cellcolor{gray!30}\textbf{1.000}(1.000)&0.667(0.000)\\ \hline
F12(2D)& \cellcolor{gray!30}\textbf{1.000}(1.000)&0.993(0.940)&0.995(0.960)&\cellcolor{gray!30}\textbf{1.000}(1.000)&0.990(0.920)&\cellcolor{gray!30}\textbf{1.000}(1.000)&0.935(0.935)&0.998(0.980)&0.710(0.000)\\ \hline
F13(2D)&\cellcolor{gray!30}\textbf{1.000}(1.000)&0.687(0.000)&0.920(0.700)&\cellcolor{gray!30}\textbf{1.000}(1.000)&0.833(0.140)&\cellcolor{gray!30}\textbf{1.000}(1.000)&0.667(0.000)&0.990(0.940)&0.667(0.000)\\ \hline
F14(3D)&\cellcolor{gray!30}\textbf{1.000}(1.000)&0.667(0.000)&0.830(0.180)&0.917(0.560)&0.710(0.000)&0.760(0.060)&0.667(0.000)&0.710(0.000)&0.667(0.000)\\ \hline
F15(3D)&0.750(0.000)&0.748(0.000)&0.675(0.000)&0.750(0.000)&0.748(0.000)&0.585(0.000)&0.618(0.000)&0.670(0.000)&0.483(0.000)\\ \hline
F16(5D)&\cellcolor{gray!30}\textbf{0.833}(0.000)&0.667(0.000)&0.673(0.000)&0.687(0.000)&0.667(0.000)&0.657(0.000)&0.650(0.000)&0.660(0.000)&0.667(0.000)\\ \hline
F17(5D)& \cellcolor{gray!30}\textbf{0.750}(0.000)&0.593(0.000)&0.653(0.000)& \cellcolor{gray!30}\textbf{0.750}(0.000)&0.700(0.000)&0.335(0.000)&0.505(0.000)&0.538(0.000)&0.180(0.000)\\ \hline
F18(10D)&\cellcolor{gray!30}\textbf{0.667}(0.000)&\cellcolor{gray!30}\textbf{0.667}(0.000)&\cellcolor{gray!30}\textbf{0.667}(0.000)&\cellcolor{gray!30}\textbf{0.667}(0.000)&\cellcolor{gray!30}\textbf{0.667}(0.000)&0.333(0.000)&0.497(0.000)&0.633(0.000)&0.510(0.000)\\ \hline
F19(10D)&\cellcolor{gray!30}\textbf{0.625}(0.000)&0.475(0.000)&0.538(0.000)&0.585(0.000)&0.510(0.000)&0.135(0.000)&0.223(0.000)&0.447(0.000)&0.000(0.000)\\ \hline
F20(20D)&\cellcolor{gray!30}\textbf{0.500}(0.000)&0.290(0.000)&0.468(0.000)&0.483(0.000)&0.480(0.000)&0.050(0.000)&0.125(0.000)&0.178(0.000)&0.000(0.000)\\ \hline
Rank& \cellcolor{gray!30}\textbf{4.9}&10.65&8.25&5.3&9.525&9.3&10.85&9.2&13.525\\\hline
\multicolumn{2}{c|}{+  /  =  /  -}&12/8/0&12/8/0&5/14/1&14/6/0&8/11/1&10/9/1&11/8/1&14/6/0\\ \hline

    \end{tabular}
     }
\end{table*}

\section{Experimental Results}
\subsection{Settings}
We adopt the well-known CEC2013 MMOP benchmark~\cite{li2013benchmark} to evaluate APDMMO, which includes $20$ functions with varying properties. Among them, F1 to F5 are simple functions, F6 to F10 are scalable functions with many global optima, and F11 to F20 are composition functions with challenging landscapes. These functions cover different dimensions and the number of global optima. 

To measure the performance of MMOP algorithms, two metrics are commonly used: Peak Ratio (PR) and Success Rate (SR). Given a certain accuracy, an optimum is claimed to be found if there exists a solution near it, whose objective value meets the accuracy requirement. And a run is defined as successful if all global optima are found within that single run. PR is calculated as the percentage of the number of found optima out of the total number of optima, while SR is calculated as the ratio of successful runs to the total number of runs. The formulas of these two metrics are:
\begin{equation}
    PR = \frac{\sum_{i=1}^{NR} NPF_i}{NKP \cdot NR}, SR = \frac{NSR}{NR}
\end{equation}
where $NR$ is the number of runs, $NPF_i$ represents the number of found optima in the $i$-th run, and $NKP$ is the total number of global optima of the problem. $NSR$ means the number of successful runs. In short, PR reflects the ability of algorithms to locate more optima, while SR represents the probability to locate all global optima.

To align the parameters' granularity with the problems' dimensions, we set them as follows. In the GLF stage, we set the train ratio ($r$) to be $3/8$. The hidden dimension ($h$) for Landscape Learner is $128$, while the number of NLA blocks, $K$, is $5$. The training epoch is $400$ with a batch size of $400$ and a learning rate of $0.0005$. The number of levels, $s$, is $10$. In the FPD stage, we use $1000000$ samples with a step size of $0.005$ . The number of optimization steps is $3000$ when the problem dimension ($D$) is less than $20$, otherwise, it is $5000$. During the process of DBSCAN,~ the parameter settings are: $eps=0.1, minPts=2 (D<5);eps=0.1,minPts=20 (5\leq D<10); eps=0.2, minPts=40 (D\geq10)$. In the PLS stage, if $D<5$, $n_{pls}$ is $8$; if $5 \leq D < 20$, $n_{pls}$ is $10$, otherwise, $n_{pls}$ is $20$. And the step size $\sigma $ is $ 0.1$ when $D < 5$ and $0.5$ when $D \geq 5$. The optimization generation  $G_{pls}$ is $200$. Additionally, the $\epsilon_{es}$ and $k_{es}$ for early stopping is $1\times10^{-5}$ and $20 \times n_{pls}$, respectively. The maximum number of function evaluations for each function adheres to the settings specified in the CEC2013 MMOP benchmark~\cite{li2013benchmark}. The experiments are conducted on a computer with an Intel(R) Core(TM) i9-10980XE CPU, 128G of RAM and a NVIDIA  RTX 3090 GPU.

\begin{figure*}[h]
	\centering
	\begin{minipage}{0.68\textwidth}
		\includegraphics[width=\textwidth]{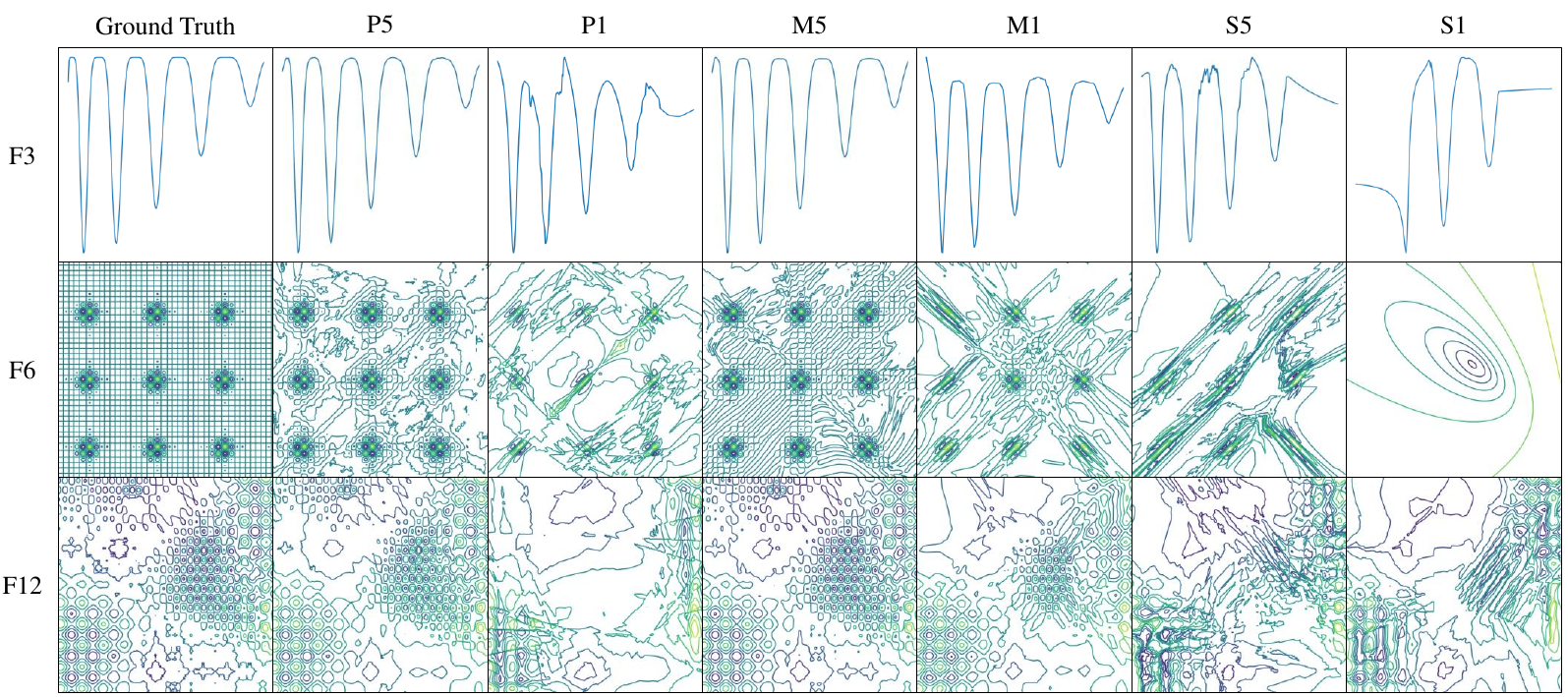}
	\end{minipage}
	\begin{minipage}{0.28\textwidth}
		\includegraphics[width=\textwidth]{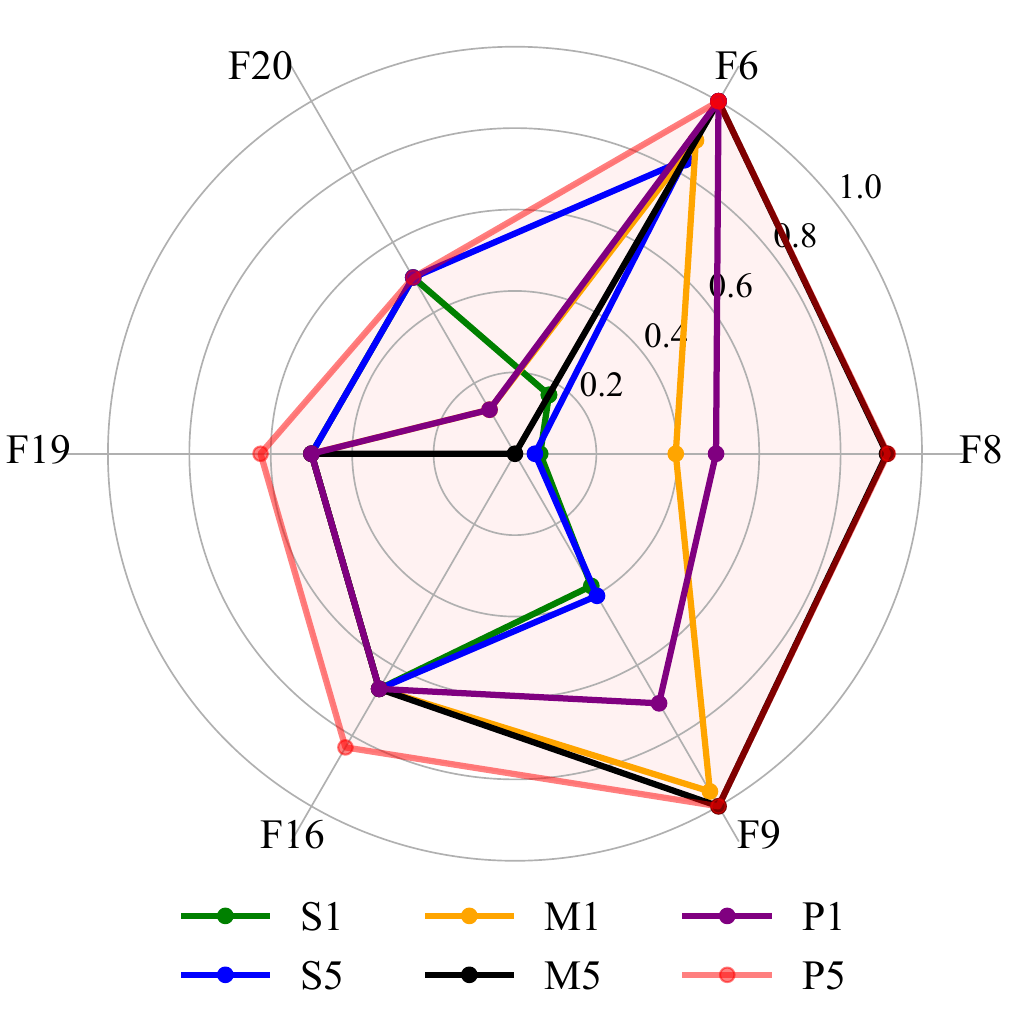}
	\end{minipage}
	
	\caption{Comparison of prediction accuracies of different Landscape Learner model architecture designs.}
	\label{pic: differentmodels}
\end{figure*}

\subsection{Performance Comparison Analysis}
We choose $16$ up-to-date MMOP algorithms as baselines to compare with our APDMMO. Specially, we select NBNC-PSO-ES~\cite{NBNC},  PMODE~\cite{wei2021penalty}, ESPDE~\cite{li2023history},  NetCDE\textsubscript{MMOPs } (shorted as NetCDE)~\cite{chen2023network}, VNCDE~\cite{zhang2019parameter}, dADE2~\cite{epitropakis2013dynamic}, CMSA-ES-DIPS~\footnote{https://github.com/fzjcdt/CMSA-ES-DIPS}, Hillvallea~\cite{hillvallea}, RS-CMSA~\cite{ahrari2017multimodal} and NMMSO~\cite{fieldsend2014running} as representative algorithms with different niching methods. And DIDE~\cite{chen2019distributed}, FBK-DE~\cite{FBK-DE}, NDC-DE~\cite{sun2023differential} are algorithms introducing new evolutionary operators. Also, we choose   EMOMMO~\cite{EMOMMO},  MOMMOP~\cite{MOMMOP} to solve MMOPs through multi-objective problem transformation. Additionally, RLEMMO~\cite{lian2024rlemmo} serves as a baseline to solve MMOPs with reinforcement learning.

All algorithms are evaluated using the CEC2013 MMOP benchmark across multiple runs, and the results are shown in Table \ref{main_exp}. PR values below an accuracy of $1\times 10^{-4}$ are selected as the primary metric, while SR values are provided in parentheses. The best results of each problem are highlighted in bold. The symbols '+', '=' and '-' indicate the number of problems on which APDMMO outperforms, performs equally to, or underperforms compared algorithms, respectively. And the ranking is determined through Friedman's test using the KEEL software~\cite{KEEL}. From the results in Table~\ref{main_exp}, we obtain several key observations and discuss them as follows.

1) APDMMO presents its effectiveness for solving MMOPs with various dimensions. For low dimensional problems ($D<5$), F1 to F15, APDMMO successfully identifies all accurate global optima in $13$ cases, including functions with large number of global optima (F9) or multiple local optima (F11-F14). This highlights that for low dimensional problems, the Landscape Learner of APDMMO can offer relatively precise and comprehensive landscape insights, enabling the discovery of all potential optima for subsequent optimization. As the dimensionality increases, the accuracy of the Landscape Learner diminishes due to the constraints of inadequate sampled data, resulting in decreased performance for high dimensional problems ($D \geq 5$). Nonetheless, APDMMO still could locate over $50\%$ of global optima precisely.

 2) Compared with other algorithms, APDMMO exhibits the best performance and high robustness. Among the $20$ problems, APDMMO gains achieves the best results in $18$ problems, demonstrating the robustness of APDMMO when facing problems with various properties. And APDMMO outperforms all of others on F14, F16, F19 and F20, showcasing the high efficiency of APDMMO for optimizing high dimensional problems. This efficiency is achieved by combining peak detection and local search to provide a good diversity and quality of solutions. According to the results of Friedman's test, APDMMO ranks first among the compared algorithms.

3) From the results, we can see that APDMMO does not perform well on F8, and slightly underperforms compared to NDC-DE on F15 ($0.750$ versus $0.753$). F8 is a $3$D Shubert function, and the distribution of objective values is so unbalanced that it is challenging for our Landscape Learner to approximate all the global peaks accurately. This limitation hinders the performance of APDMMO on such challenging scenarios. Regarding F15, we observe that for an accuracy of $1\times10^{-5}$ (the results are omitted here due to the limited space), APDMMO gains a better result than NDC-DE ($0.750$ versus $0.720$), showcasing the superiority of APDMMO over NDC-DE. 

4) APDMMO exhibits a strong competitiveness compared with algorithms in other branches of MMOPs. For niching-based algorithms without landscape knowledge, like VNCDE and NetCDE, they define neighborhoods or clusters based on the distance between individuals without considering objective values. Therefore, they cannot discover basins of peaks efficiently. EMOMMO and MOMMOP are multi-objective transformation methods, archiving good results on low dimensional problems. However, their performance drops rapidly as the dimension increases, due to the difficulty in locating Pareto front. Additionally, RL-based methods, like RLEMMO, face challenges in balancing exploration and exploitation in high dimensional problems. Compared to these algorithms, APDMMO can locate the potential  peaks efficiently through landscape approximation and peak detection, followed by local search to improve the quality of solutions. Consequently, APDMMO can solve problems with many optima or high dimensions effectively.

5) The Landscape Learner of APDMMO can acquire more comprehensive knowledge compared to the analysis methods in existing landscape-assisted MMOP works. Specially, EMOMMO approximates the landscape through locating the Pareto front, while Hillvallea and NBNC-PSO-ES reflect the landscape information based on the relationship of distances and objective values. However, these methods rely on the experience of EC optimization, making it challenging for them to detect peaks in unexplored regions or high dimensional spaces. In contrast, APDMMO analyzes the landscape by training a Landscape Learner, which can offer comprehensive  knowledge to enhance the performance of APDMMO on problems with numerous global optima (F9) and high dimensions (F19, F20).

\subsection{In-depth Discussion}

To verify the effectiveness of core designs in APDMMO, we design several ablation studies for an in-depth discussion. All experiments are conducted on the CEC2013 benchmark, and the PR(SR) at $1\times 10^{-4}$ accuracy is selected as metrics.

\subsubsection{Architecture Design Choices}
In APDMMO, the core component of the Landscape Learner we designed for surrogate learning is the NLA blocks~(Figure~\ref{fig:network}), which ensemble many activation units with diverse non-linearity to enhance the fitting accuracy on MMOPs with complex local/global landscape structures. To verify the expressiveness of NLA block, we compare it with two other architectures: 1) MLP block, which contains two linear layers with parameters in $\mathbb{R}^{500 \times 500}$, each followed by a ReLU function; 2) Sequential version of NLA block, which arranges the activation units in Figure~\ref{fig:network} in vertical series. Based on these three kinds of blocks, we test the optimization performance of six variants of APDMMO on $20$ CEC 2013 problems, which are with different architectures of the Landscape Learner: 1) P1: Using $1$ NLA block; 2) P5: Stacking $5$ NLA blocks (the same as APDMMO in this paper); 3) M1: Using $1$ MLP block; 4) M5: Stacking $5$ MLP blocks; 5) S1: Using $1$ Sequential NLA block; 6) S5: Stacking $5$ Sequential NLA blocks. We present some interesting experimental results in Figure~\ref{pic: differentmodels}. In the left of this figure, we illustrate the contour maps of some $1$D/$2$D problems to give an intuitive reflection of the learning ability of different architectures. The right of this figure shows the optimization performance of these six variants on challenging or high dimensional problems. The results show three key observations:  
 
1) Upon comparing the contour maps, it is evident that P5, M5, and S5 achieve higher accuracies than their counterparts (P1, M1 and S1 respectively), indicating that adding more parameters to a basic model~(make it deeper in our case) could enhance the landscape approximation accuracy of the surrogate model. This is consistent with the observations in classic regression researches. 

2) From the comparison between P5 (P1) and S5 (S1), we can find out that models that arrange different activation units in a multi-head form (as in our NLA block) show more vivid expressive power and landscape prediction accuracy than models that arrange these activation units in a sequential vertical way. One possible explanation behind is that: in sequential mode, different activation units comply together in a cascading manner, later activation functions may blend the effects of the previous ones, thus weakening the overall effectiveness.

3) In low dimensional problems F3, F6 and F12, the Landscape Learner models with MLP block M5 and M1 show  comparable landscape approximation accuracies with the Landscape Learner models equipped with NLA block P5~(the setting in our APDMMO) and P1. On the contrary, in high dimensional problems such as F16,F19 and F20~(right of Figure~\ref{pic: differentmodels}), APDMMO with P5 as Landscape Learner shows clear superiority than APDMMO with M5. Such observation might originate from the two aspects: On the one hand, landscape in low-dimensional problems is relatively easier to fit for both NLA block and MLP block. On the other hand, the ensemble of diverse non-linear activation units in NLA block introduces various local non-linearity, which provides more accurate landscape fitting for high-dimensional problems with various local landscapes.



\begin{figure}[t]
  \centering
  \includegraphics[width=0.7\columnwidth]{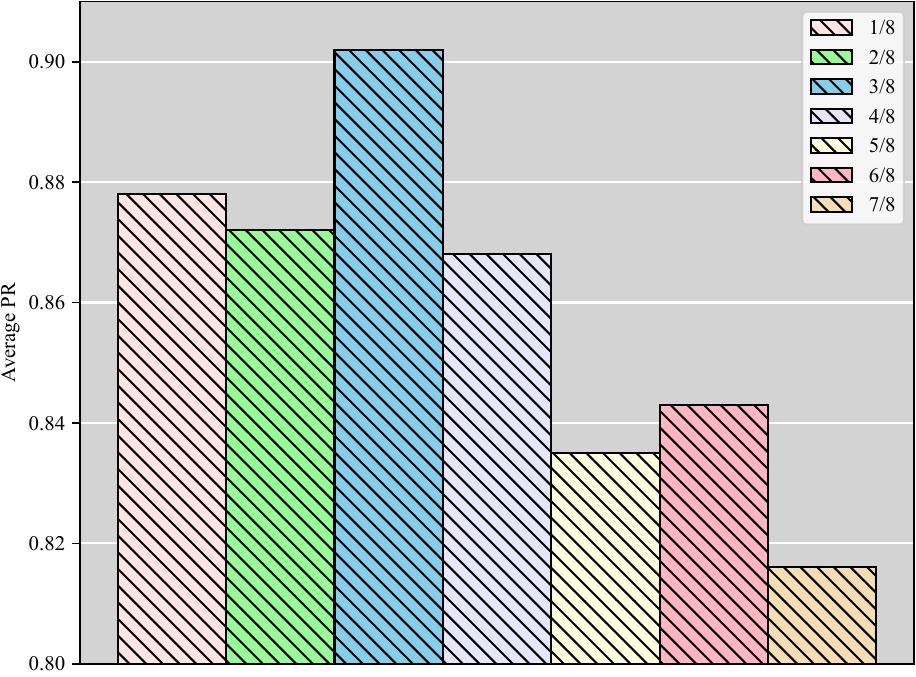}
  \vspace{-3mm}
  \caption{Average peak ratio over $f_1 \sim f_{20}$ achieved by different function evaluations allocation settings.}
  \label{fig:allocation}
  \vspace{-3mm}
\end{figure}

\subsubsection{Computational Resource Allocation~\label{res-ablation}}

The allocation of computational resource (available function evaluations) for GLF and PLS poses a tradeoff of exploration and exploitation in APDMMO. Allocating more resource to GLF can improve the accuracy of landscape approximation hence ensuring accurate peak detection, while allocating more resource to PLS can enhance the local search quality. Here we design $7$ experiments to discuss the ratio of maximal function evaluations for training, ranging from $1/8$ to $7/8$. From the average PR in figure~\ref{fig:allocation}, we can see that using $3/8$ of the maximal function evaluations to train the Landscape Learner and the remaining to process local search can achieve the best performance.

\subsubsection{Algorithm Stage Ablation}
\begin{figure}[t]
  \centering
  \includegraphics[width=0.9\columnwidth]{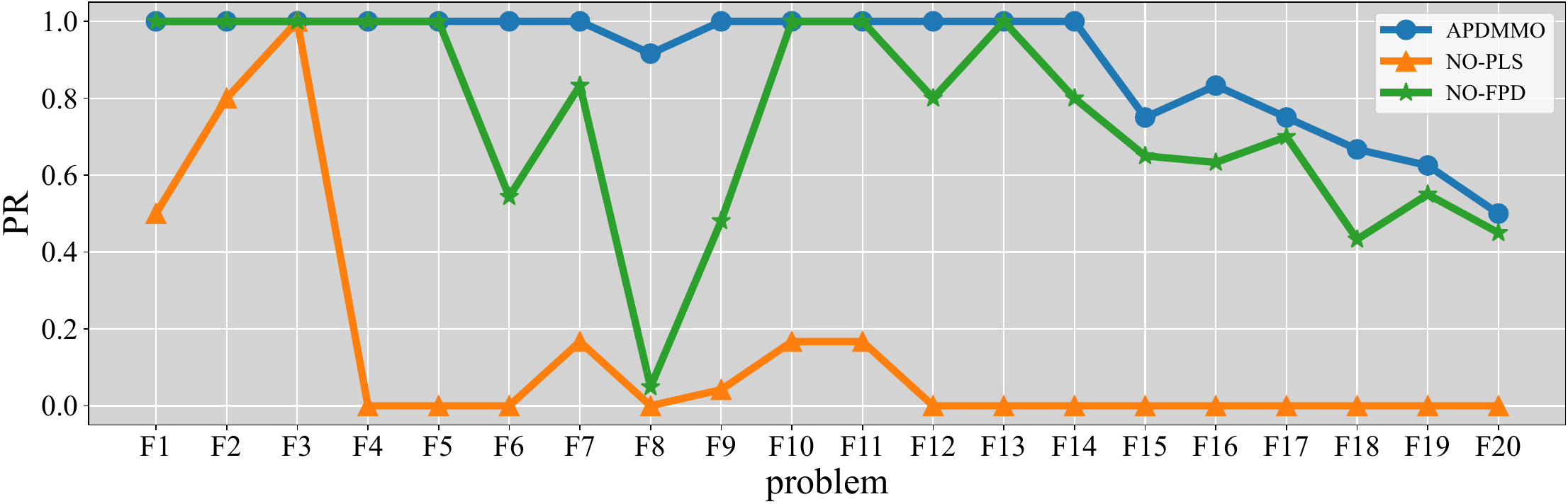}
  \vspace{-3mm}
  \caption{Importance analysis of FPD and PLS stages.}
  \label{fig:stage}
  \vspace{-4mm}
\end{figure}

In APDMMO, FPD stage detects the potential peak areas and PLS stage search for accurate optima in these areas. To verify the importance of FPD and PLS stage, we compare APDMMO with two variants NO-FPD and NO-PLS. The former directly uses SEP-CMAES to perform local search for many randomly sampled areas. The latter only goes through GLF stage and FPD stage. We present the PR of APDMMO and these two variants on all 20 tested problems in Figure~\ref{fig:stage}. Obviously, NO-PLS fails to locate global optima in many problems, due to the prediction bias in the trained Landscape Learner. In contrast, NO-FPD performs worse than APDMMO on problems with challenging landscapes (e.g., F6, F8), multiple global optima (e.g., F7 and F9) or high dimensions (e.g., F15 to F20), since there is no guidance for the starting positions for local search. Therefore, the FPD and PLS are complementary stages: FPD provides starting positions near optima to enhance the efficiency of local search, while PLS refines potential optima to improve the accuracy of results. These two stages contribute to APDMMO's robustness on tested benchmark.

\section{Conclusion}

In this paper, we propose APDMMO as a novel optimization framework for solving MMOPs. We design a novel network architecture to ensemble diverse non-linear activation units for accurate landscape learning, and propose an efficient multi-start adamW strategy assisted by the landscape surrogate for peak detection. After locating the potential peaks, we employ multiple SEP-CMAES instances for local search. The experimental results demonstrate the superiority of APDMMO over several up-to-date baselines and the important roles of the core designs of APDMMO. We would like to note that APDMMO represents a novel exploration of combining the strength of efficient gradient-based optimization and effective EC optimization. In this paper, we demonstrate learning a surrogate model could be a promising way to bridge these two optimization techniques. 


\begin{acks}
This work was supported in part by the National Natural Science Foundation of China No. 62276100, in part by the Guangdong Provincial Natural Science Foundation for Outstanding Youth Team Project No. 2024B1515040010, in part by the Guangdong Natural Science Funds for Distinguished Young Scholars No. 2022B1515020049, and in part by the TCL Young Scholars Program.
\end{acks}

\bibliographystyle{ACM-Reference-Format}
\bibliography{myref}


\end{document}